# Signed Spiking Neuron Enabled by an Orthogonal-Easy-Axis Magnetic Tunnel Junction

Huannan Zheng [1], Jingli Liu [1], Kezhou Yang [1, *]

[1] Function Hub, The Hong Kong University of Science and Technology (Guangzhou), Guangzhou, 51000, China
[*] Corresponding author, email: kezhouyang@hkust-gz.edu.cn

**Abstract—**Signed spiking neurons carry richer information than standard spiking neurons. This work proposes a compact magnetic tunnel junction (MTJ)-based neuron for signed leaky integrate-and-fire (LIF) operation. With orthogonal easy axes in the free and pinned layers, the device enables bipolar spike generation and maps magnetic-moment dynamics to signed LIF membrane-potential evolution. Landau-Lifshitz-Gilbert simulations show that proper free-layer dimensions allow the device response to follow a signed LIF equation. A representative design of 10 nm × 45 nm × 50 nm corresponds to an aspect ratio of about 2:9:10. Network evaluations using the fitted device-neuron model achieve 91.06% on CIFAR-10 and 77.40% on CIFAR10-DVS, retaining most of the accuracy of ideal signed LIF neurons.

## I. INTRODUCTION

Emerging device technologies have been widely investigated to mitigate data-movement-induced energy and latency overheads in von Neumann computing systems. Among them, spintronic devices offer a promising route toward compact and energy-efficient neuromorphic hardware [1], [2]. Magnetic tunnel junctions (MTJs), in particular, have been explored for implementing leaky-integrate-and-fire (LIF) neurons [3], [4] and synaptic functions [4], [5]. However, most reported MTJ-based neuron implementations focus mainly on conventional LIF model, which makes signed information representation inefficient over short temporal windows [6]-[8]. At the algorithmic level, signed LIF neurons have been proposed to improve information representation by emitting positive and negative spikes when the membrane potential reaches the corresponding thresholds [9]-[11]. Despite this advantage, compact device-level implementations of signed LIF dynamics remain underexplored.

In this work, we propose an MTJ-based signed LIF device neuron by introducing orthogonal easy-axis directions in the free and pinned layers. The pinned-layer easy axis is aligned with the short-axis direction of the elliptical free layer, enabling bidirectional magnetic-moment accumulation under input currents of opposite polarities. The simulated device dynamics are consistent with signed LIF membrane-potential evolution and are incorporated into network-level evaluations through a fitted device-neuron model.

## II. RESULTS AND DISCUSSION

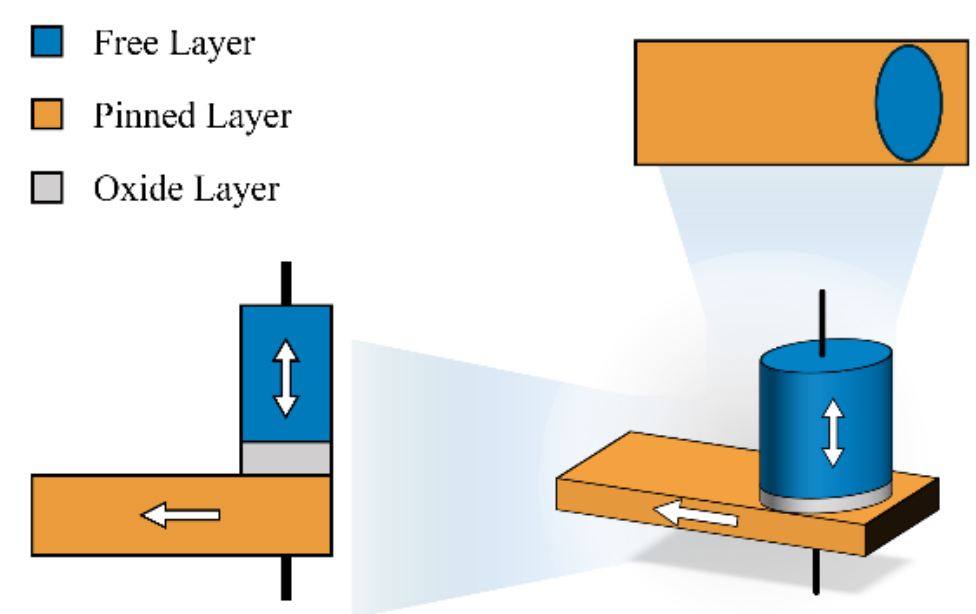

Fig. 1. Schematic structure of the proposed MTJ-based device

Table I.
DEVICE AND SIMULATION PARAMETERS

| Parameter | Symbol | Value | Unit |
|---|---|---|---|
| Short axis of free layer | $W$ | 10 | nm |
| Long axis of free layer | $L$ | 45 | nm |
| Free-layer thickness | $t$ | 50 | nm |
| Saturation magnetization [12] | $M_s$ | $1.15 \times 10^6$ | A/m |
| Gilbert damping coefficient [13] | $\alpha_G$ | 0.01 | - |
| Spin efficiency [14] | $\eta$ | 0.5 | - |
| Temperature | $T$ | 300 | K |

The proposed device structure is schematically illustrated in Fig. 1. The free layer and pinned layer are magnetic layers separated by an oxide spacer layer. The pillar-shaped free layer has an easy axis along the height direction, whereas the pinned-layer easy axis lies along the in-plane short-axis direction of the free layer. The proposed MTJ is a two-terminal device driven by an input current through the free and pinned layers. The current is spin-polarized by the pinned layer and then exerts an in-plane spin-transfer torque on the free layer. This torque drives the free-layer magnetization toward one of two opposite in-plane directions, depending on the current polarity.

错误!未找到引用源。 shows the magnetic-moment dynamics under current pulses. In 错误!未找到引用源。(a), the LLG-simulated trajectory of $\mathbf{m}$ shows that opposite current polarities drive the free-layer magnetization toward two opposite in-plane directions. The simulation parameters are listed in Table 1. 错误!未找到引用源。 (b) shows the projection of $\mathbf{m}$ along the pinned-layer easy axis, indicating bidirectional integration and leakage for signed LIF operation. The MTJ resistance, determined by the relative angle between the free and pinned layers, is converted to $V_{out}$ using the voltage divider in 错误!未找到引用源。 (c). In 错误!未找到引用源。 (d), spikes are triggered when the equivalent membrane potential reaches the positive or negative threshold. The free-layer dynamics are modeled by the LLG equation with spin-transfer torque [16]:

$$\frac{d\mathbf{m}}{dt} = -\gamma\mathbf{m} \times \mathbf{H}_{\text{eff}} + \alpha_G \mathbf{m} \times \frac{d\mathbf{m}}{dt} + \frac{1}{qN_s}(\mathbf{m} \times \mathbf{I}_s \times \mathbf{m}), \quad (1)$$

where $\gamma$ is the gyromagnetic ratio, $\alpha_G$ is the Gilbert damping coefficient, $\mathbf{H}_{\text{eff}}$ is the effective magnetic field, $N_s = M_s V/\mu_B$ is the number of spins in free layer of volume $V$ ($M_s$ is saturation magnetization and $\mu_B$ is Bohr magneton), $q$ represents the the elementary charge, $\mathbf{I}_s$ is the input spin current.

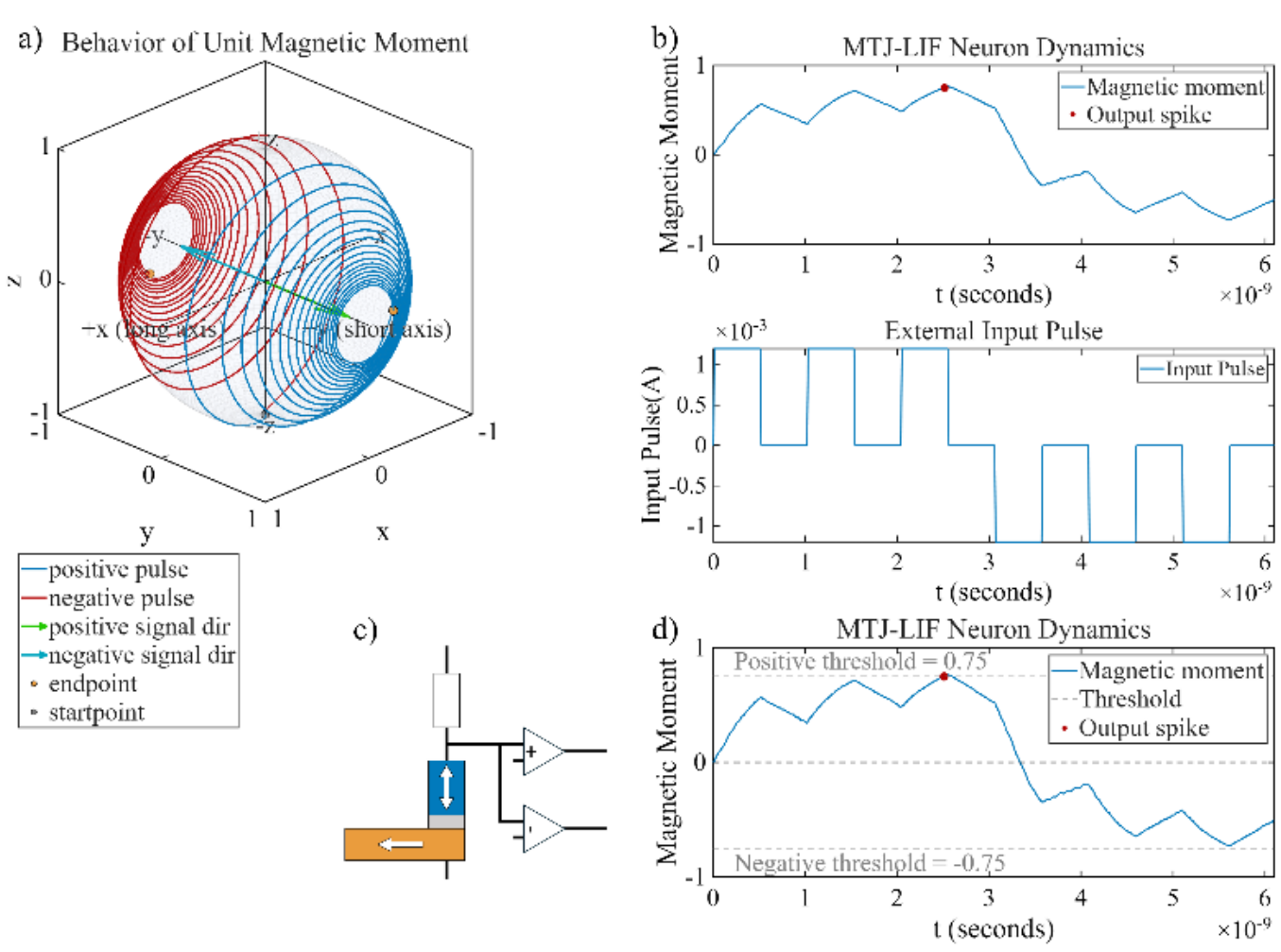

Fig. 2. a) Behavior of the unit magnetic moment of the free layer under current pulses with opposite polarities b) Evolution of the free-layer magnetic moment component along the pinned-layer easy-axis direction under an input pulse sequence c) Circuit design for bipolar spike generation based on the proposed device d) Equivalent membrane-potential dynamics with positive and negative thresholds

Fig. 3 (a) compares the state evolution of the ideal LIF model, the proposed orthogonal MTJ, and the conventional parallel MTJ [15], [16]. The parallel MTJ deviates strongly from LIF dynamics, making the corresponding networks difficult to train. In contrast, the orthogonal design better matches the LIF membrane-potential behavior. Its LLG-simulated trajectory can therefore be fitted by the LIF equation with constant parameters, as shown in Eq. (2):

$$\tau_m \frac{du}{dt} = -\alpha_1 u + \alpha_2 I_{\text{in}} R_m, \quad (2)$$

where $u$ denotes the equivalent membrane potential, corresponding to the magnetization projection along the pinned-layer easy axis. $\tau_m$ is the membrane time constant, $\alpha_1$ and $\alpha_2$ characterize leakage and input integration, respectively. $I_{\text{in}}$ is the applied input current, and $R_m$ is the effective input resistance or scaling factor in the fitted LIF model.

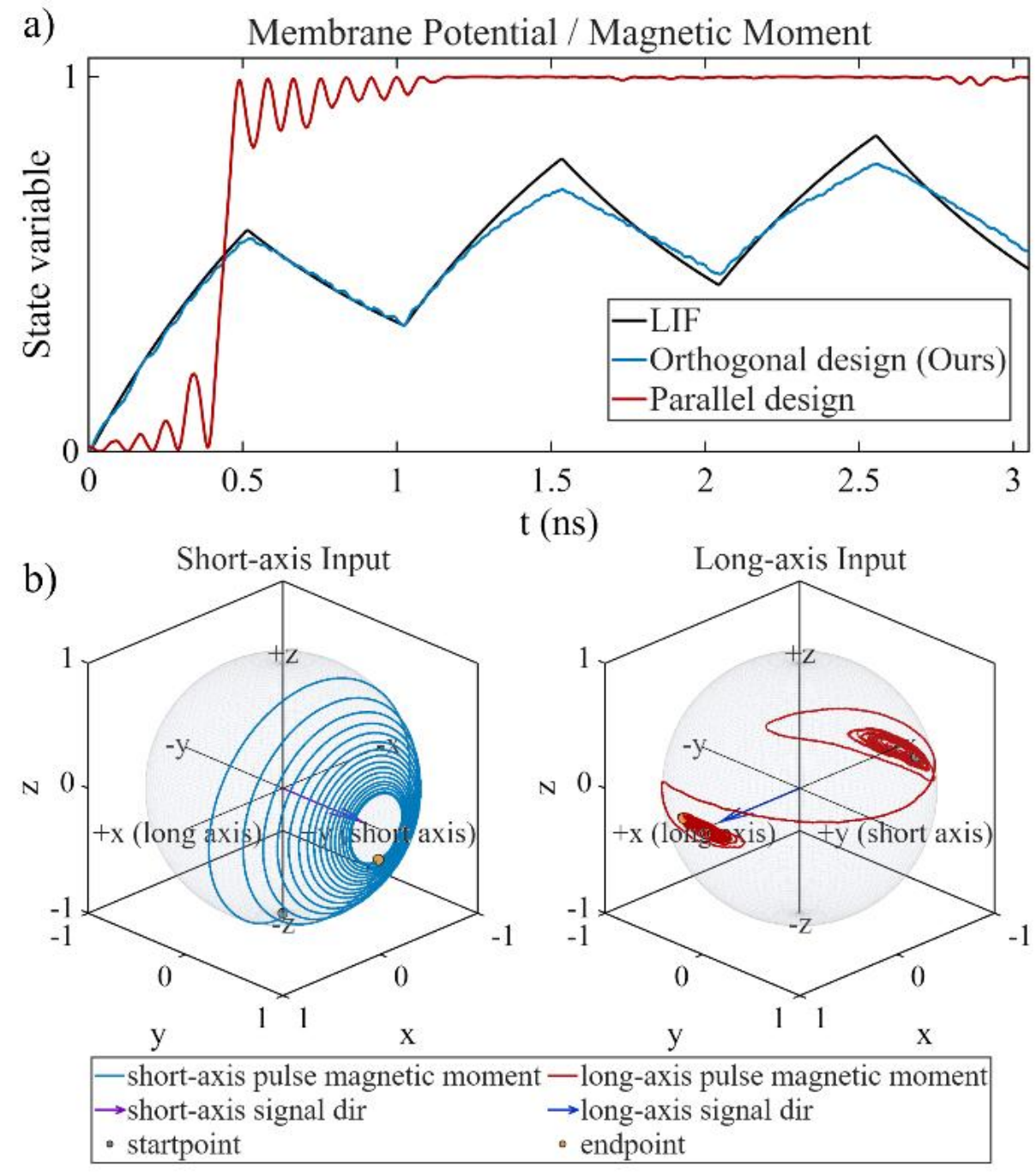

Fig. 3. a) Comparison of the state evolution among the ideal LIF neuron, the proposed device, and the conventional MTJ-based design. b) Magnetic-moment trajectories under short-axis input and long-axis input, showing the different dynamic behaviors induced by the two configurations.

The physical origin of this difference is illustrated in Fig. 3 (b). In the proposed design, the easy axis of the pinned layer is aligned with the short axis of the elliptical free layer, enabling the free-layer magnetic moment to precess along the short-axis direction under current input. This configuration produces gradual accumulation and relaxation similar to the LIF membrane-potential dynamics, whereas the conventional long-axis-input configuration fails to reproduce the desired LIF-like behavior.

To further analyze the free-layer magnetic-moment motion, the LLG equation in Eq. (1) is examined. The dynamic equation of the short-axis component of the free-layer magnetic moment, $m_y$, can be derived as

$$\begin{aligned}\frac{dm_y}{d\tau} &= (H_z - H_x)m_x m_z \\ &+\alpha_G m_y\left[(H_y - H_x)m_x^2 + (H_y - H_z)m_z^2\right] \\ &+s(t)\left(1 - m_y^2\right),\end{aligned} \tag{3}$$

In Eq. (3), the three terms describe precession, Gilbert damping, and input-driven spin torque, respectively. Spin torque induces state accumulation, while damping provides relaxation. The key difference between the orthogonal and parallel designs is the damping-related term containing $H_y - H_x$. Owing to the relative values of $H_y$ and $H_x$, this term has an opposite sign in the two designs. This gives the orthogonal MTJ a damping behavior that better resembles the LIF membrane potential. Since $H_x$ and $H_y$ depend on the free-layer dimensions, they determine the balance between accumulation and leakage. Proper dimensions are therefore required for controllable short-axis motion and LIF-like dynamics.

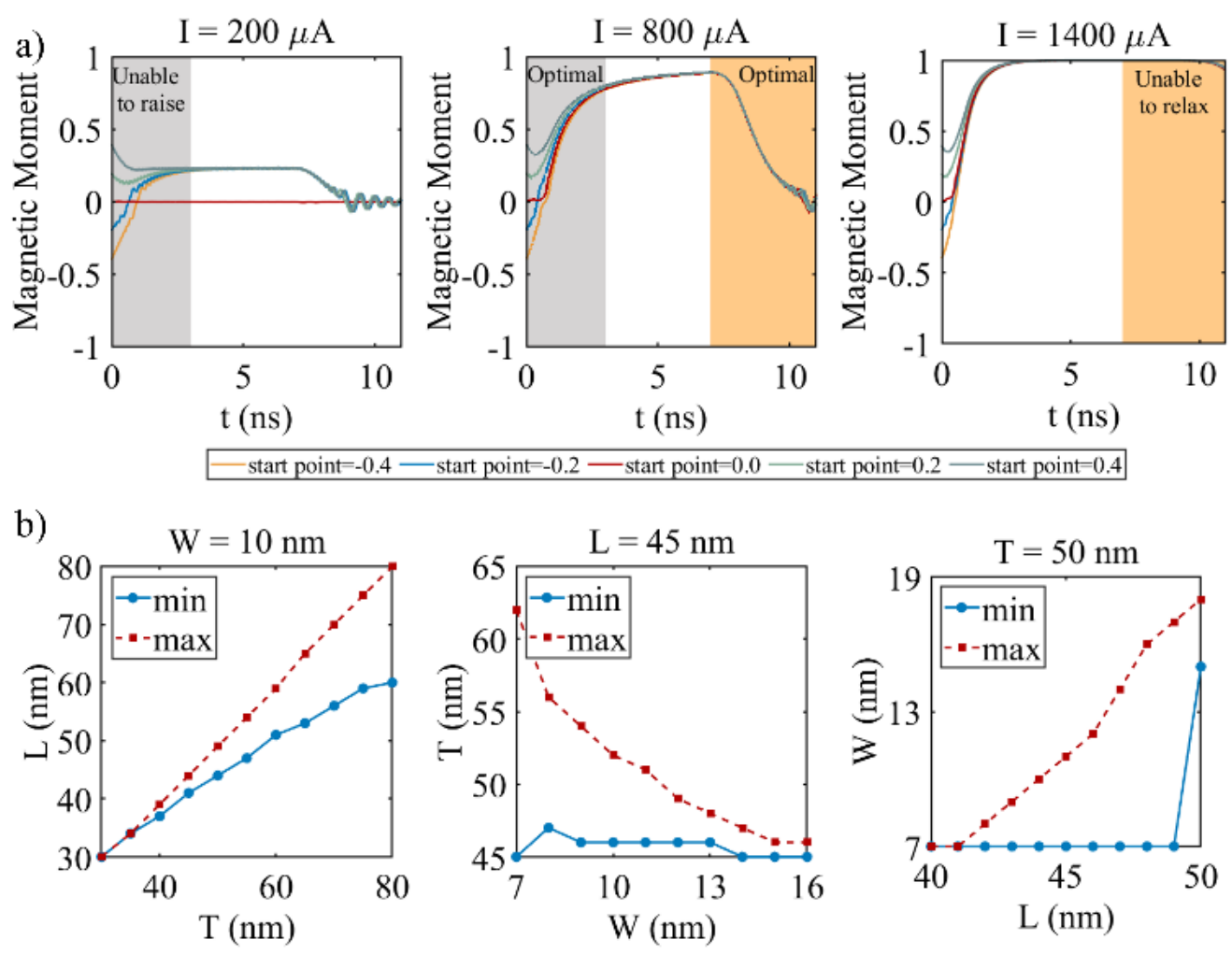

Fig. 4. a) Representative magnetization evolution under different operating conditions. b) Extracted dimensional design window of the proposed device.

The effect of device dimensions on magnetization dynamics is further examined. Fig. 4(a) shows three representative regimes. With proper matching between input current and device dimensions, the magnetization can reach the threshold and relax after input removal, enabling stable integration, thresholding, and leakage. Fig. 4 (b) summarizes the extracted dimension window. The allowable ranges show an approximately linear dependence when one dimension is fixed and the other two are swept, indicating a feasible design window for signed LIF operation. A representative window is found around an aspect ratio of $W:L:T = 2:9:10$.

Table II.
NETWORK PERFORMANCE OF THE PROPOSED DEVICE NEURON

| Dataset | Method | Accuracy |
|---|---|---|
| CIFAR-10 | ResNet18 (baseline) [17] | 93.02% |
| | Device-ResNet18 | 91.06% |
| CIFAR-10-DVS | Spikformer (baseline) [18] | 80.90% |
| | Deviceformer | 77.40% |

To evaluate system-level performance, the ideal signed LIF neurons in the baseline networks were replaced by the fitted device-neuron model, with the architectures and training/evaluation settings unchanged. As shown in Table II, Device-ResNet18 achieves 91.06% accuracy on CIFAR-10, compared with 93.02% for the reproduced baseline ResNet18 [17]. On CIFAR10-DVS, Deviceformer achieves 77.40%, compared with 80.90% for Spikformer [18]. The accuracy drops are 1.96% and 3.50%, respectively. These results show that the proposed device-neuron model preserves most of the classification performance of the ideal neuron model. The remaining loss mainly comes from the fitting error between the magnetic-moment dynamics and the ideal signed LIF equation. Thus, the proposed device neuron offers a compact hardware-compatible implementation of signed LIF dynamics while maintaining competitive performance on frame-based and event-based vision tasks.

## III. CONCLUSION

This work proposes a compact MTJ-based device neuron for signed LIF operation. Enabled by shape anisotropy, the orthogonal easy-axis configuration supports bipolar state evolution and spike generation. LLG analysis verifies that the short-axis magnetization motion can reproduce signed LIF membrane-potential dynamics, with a representative dimension window near $W\colon L\colon T = 2\colon 9\colon 10$. Network simulations show that the fitted device-neuron model achieves 91.06% on CIFAR-10 and 77.40% on CIFAR10-DVS, with only moderate accuracy degradation from the ideal-neuron baseline.

## REFERENCES

[1] K. Roy, et al. “Spintronic neural systems.” Nat Rev Electr Eng vol. 1, pp. 714–729, 2024. DOI: 10.1038/s44287-024-00107-9

[2] S. Jung, et al. “A crossbar array of magnetoresistive memory devices for in-memory computing.” Nature vol. 601, pp. 211–216, 2022. DOI: 10.1038/s41586-021-04196-6.

[3] L. Liu, et al. “Domain wall magnetic tunnel junction-based artificial synapses and neurons for all-spin neuromorphic hardware.” Nat. Commun. vol. 15, no. 4534, 2024. DOI: doi.org/10.1038/s41467-024-48631-4.

[4] Z. Chen, et al. “Nanoscale exchange-bias magnetic tunnel junctions enabled memristive synapse and leaky-integrate-fire neuron for neuromorphic computing.” Nat. Commun. Mar. 24, 2026, to be published. DOI: 10.1038/s41467-026-70802-8. 50

[5] P. Zhou, et al. “Neuromorphic Hebbian learning with magnetic tunnel junction synapses.” Commun. Eng. vol. 4, pp. 142, 2025. DOI: 10.1038/s44172-025-00479-2.

[6] L. Farcis, et al. “Spiking dynamics in dual free layer perpendicular magnetic tunnel junctions,” Nano Letters, vol. 23, no. 17, pp. 7869–7875, 2023, DOI: 10.1021/acs.nanolett.3c01597.

[7] W. H. Brigner, J. Alsaleem, K. Zahedinejad, J. A. C. Incorvia, and S. Rakheja, “Domain Wall Leaky Integrate-and-Fire Neurons with Shape-Based Configurable Activation Functions,” IEEE T. Electron Dev., vol. 69, no. 5, pp. 2353–2359, May 2022. doi: 10.1109/TED.2022.3159508. [Online]. Available: https://doi.org/10.1109/TED.2022.3159508 61

[8] T. Leonard, N. Zogbi, S. Liu, W. S. Rogers, C. H. Bennett, and J. A. C. Incorvia, “Shape Anisotropy-Dependent Leaking in Magnetic Neurons for Bio-Mimetic Neuromorphic Computing,” ACS Nano, vol. 19, no. 3, pp. 3470–3477, 2025. doi: 10.1021/acsnano.4c13020. [Online]. Available: https://doi.org/10.1021/acsnano.4c13020

[9] D. Zhao, Y. Zeng, and Y. Li, “BackEISNN: A Deep Spiking Neural Network With Adaptive Self-Feedback and Balanced Excitatory–Inhibitory Neurons,” Neural Networks, vol. 154, pp. 68–77, 2022. doi: 10.1016/j.neunet.2022.06.036. [Online]. Available: https://doi.org/10.1016/j.neunet.2022.06.036

[10] C. Zou, X. Cui, S. Feng, G. Chen, Y. Zhong, Z. Dai, and Y. Wang, “An All Integer-Based Spiking Neural Network With Dynamic Threshold Adaptation,” Frontiers in Neuroscience, vol. 18, Art. no. 1449020, 2024. doi: 10.3389/fnins.2024.1449020. [Online]. Available: https://doi.org/10.3389/fnins.2024.1449020 76

[11] Y. Hu, et al. “Fast-snn: Fast spiking neural network by converting quantized ann.” IEEE Trans. Pattern Anal. Mach. Intell. vol. 45, no. 12, pp. 14546-14562, Dec. 2023, DOI: 10.1109/TPAMI.2023.3275769.

[12] R. Heindl, et al. “Size dependence of intrinsic spin transfer switching current density in elliptical spin valves.” Appl. Phys. Lett. vol. 92, no. 26, 2008. DOI: 10.1063/1.2953980.

[13] S. Li, et al. “Magnetization Dynamics Modulated by Dzyaloshinskii-Moriya Interaction in the Double-Interface Spin-Transfer Torque Magnetic Tunnel Junction.” Nanoscale Research Letters vol. 14, no. 315, 2019. DOI: 10.1186/s11671-019-3150-4. 86

[14] A. Sengupta, et al. “Magnetic Tunnel Junction Mimics Stochastic Cortical Spiking Neurons.” Scientific Reports vol. 6, no. 30039, 2016. DOI: 10.1038/srep30039.

[15] C.-T. Tung, et al. “A Compact Model of Perpendicular Spin-Transfer-Torque Magnetic Tunnel Junction,” IEEE T. Electron Dev., vol. 71, no. 1, pp. 57–61, Jan. 2024, doi: 10.1109/TED.2023.3313997.

[16] A. Meo, et al. “Spin Transfer Torque Switching Dynamics in CoFeB/MgO Magnetic Tunnel Junctions.”

Physical Review B, vol. 103, no. 5, Art. no. 054426, 2021. DOI: 10.1103/PhysRevB.103.054426.

[17] K. He, et al. “Deep Residual Learning for Image Recognition.” Proc. IEEE Conference on Computer Vision and Pattern Recognition (CVPR), pp. 770–778, 2016. DOI: 10.1109/CVPR.2016.90.

[18] Z. Zhou, et al. “Spikformer: When Spiking Neural Network Meets Transformer.” Proc. International Conference on Learning Representations (ICLR), 2023.